\documentclass{article}

\PassOptionsToPackage{numbers, compress}{natbib}
\usepackage[dblblindworkshop, final]{neurips_2026}

\usepackage[utf8]{inputenc}
\usepackage[T1]{fontenc}
\usepackage{hyperref}
\usepackage{url}
\usepackage{booktabs}
\usepackage{amsfonts}
\usepackage{nicefrac}
\usepackage{microtype}
\usepackage{graphicx}
\usepackage{xcolor}
\usepackage{amsmath}
\usepackage{multirow}

\makeatletter
\renewcommand{\section}{\@startsection{section}{1}{\z@}%
  {-1.4ex \@plus -0.4ex \@minus -0.2ex}%
  { 1.0ex \@plus  0.2ex \@minus  0.1ex}%
  {\large\bf\raggedright}}
\renewcommand{\paragraph}{\@startsection{paragraph}{4}{\z@}%
  {1.0ex \@plus 0.3ex \@minus 0.2ex}%
  {-1em}%
  {\normalsize\bf}}
\makeatother

\workshoptitle{TAE (Trust-AI-Eval): Can We Trust AI Evaluation?}

\title{Policy Is Not Capability:\\Some Safety Benchmarks Measure Which Definition of ``Unsafe'' You Trained On}

\author{%
  Vahid Jebraeeli \quad Jonathon Romero \quad Yijin Wang \quad Adrian Chan \quad Jianbang Zhang \\
  Lenovo \\
  \texttt{vjebraeeli@lenovo.com} \\
}

\begin{document}

\maketitle

\begin{abstract}
A guardrail classifier is reported as if \emph{unsafe} were a property of text. It is not: it is a
property of a policy, and the policy enters the model through its labels. We exploit an accident of
an industrial deployment in which the same ${\sim}53$k prompts were labelled twice, under two
operational definitions (``is this a request for harmful content?'' and ``would a moderator block
this?''), and used to train ten guardrail arms sharing architecture, codebase and optimiser family,
at two model scales.
First, the guardrail's own training corpus already contained the benchmarks: JailbreakBench-benign is
$88\%$ and OR-Bench-hard $82\%$ present in the training pools, two \emph{over-refusal}
benchmarks, the kind a guardrail reports to argue it does not over-block. Everything below is
decontaminated.
Second, safety benchmarks stratify sharply by how much of their score is decided by the label
definition rather than by the model: with architecture and optimiser fixed, the range across the ten
arms is $0.99$ on HarmBench-copyright and $0.043$ on WildGuard-benign, a $23\times$ spread.
Third, a benchmark score can be moved across its entire range by $0.03\%$ of the parameters. Across
these arms HarmBench-copyright spans $0$ to $100$ and OR-Bench-hard over-refusal $3.4\%$ to $56.4\%$,
and the four arms whose \emph{last} supervision signal carries a given policy occupy one extreme of
both (exact rank $p=0.005$, or $0.04$--$0.14$ at worst), at unchanged detection capability. For a
distilled student the teacher's soft labels do this even when every hard label carries the other
policy, so teacher provenance is an unrecorded confound in any released checkpoint.
Fourth, validation accuracy on the one split that grades all twenty arms tracks agreement with the
annotator's policy ($\rho = +0.74$), not detection capability ($\rho = +0.34$).
We argue that guardrail evaluations should report contamination, a seed floor, and the definition
they were graded against as first-class variables.
\end{abstract}

\section{Introduction}

Safety classifiers, commonly called guardrails, are trained and reported like any other classifier.
A paper states a macro-F1 on a held-out split, a set of catch rates on public benchmarks such as
HarmBench~\citep{mazeika2024harmbench}, AdvBench~\citep{zou2023advbench},
JailbreakBench~\citep{chao2024jailbreakbench} or StrongREJECT~\citep{souly2024strongreject},
an over-refusal number on OR-Bench~\citep{cui2024orbench} or XSTest~\citep{rottger2024xstest},
and the reader is invited to compare that row against another system's row.

This reporting convention contains a silent assumption: that the label \emph{unsafe} names a
property of the prompt, so that two systems scored on the same prompts are being compared on the
same question. In deployment that assumption is visibly false. ``Reproduce the lyrics of this song''
is a policy violation for a content platform and an ordinary request for a drafting assistant.
``Tell me a dirty joke'' is blocked by a moderation product and answered by a coding assistant.
Neither is a fact about the string. Each is a decision, made once, by whoever wrote the labelling
policy, and then absorbed by every model trained on those labels.

Most work cannot separate the two, because a project has one labelling policy and therefore one
model family. We separated them by accident. Building a guardrail against a proprietary 24-pillar
safety taxonomy for an on-device assistant, we labelled the same prompt pool twice, a month apart,
with two pipelines judged comparably accurate at the time. They disagreed on one prompt in six, and
the disagreements were not noise: they were systematic, and traced to a difference in the question
the pipeline was answering.
Pipeline~A asked \emph{would a thoughtful moderator block this?} Pipeline~B asked
\emph{is this a request for harmful content?}
We then trained ten arms (mixing, re-weighting, anchoring, distilling and head-calibrating
across the two label families) at two model scales, and evaluated all twenty models identically on
fifteen public safety benchmarks ($33{,}137$ prompts).

That grid is a measuring instrument for the evaluation protocols themselves. Our contributions
are diagnostic rather than methodological:

\begin{enumerate}\itemsep0pt\parskip0pt
\item \textbf{The training corpus contains the benchmark suite} (\S\ref{sec:contam}). We measure the
overlap between every benchmark and every pool the grid trained or validated on. It is severe and
structured: $88\%$ of JailbreakBench-benign and $82\%$ of OR-Bench-hard, against $0\%$ of
OR-Bench-toxic and $2\%$ of AdvBench. The worst hit are the over-refusal benchmarks, because the
training class they resemble was built by harvesting them. Everything below is decontaminated.
\item \textbf{Benchmarks differ enormously in how much of their score the label definition decides}
(\S\ref{sec:sens}). With architecture and optimiser fixed, changing where the two label families
enter training moves HarmBench-copyright across its entire range ($0.99$ of $1.0$) and
WildGuard-benign hardly at all ($0.043$). But a definition-unstable benchmark is also seed-unstable,
and separating the two needs a seed floor almost nobody reports; we measure one from this codebase
and show it leaves the middle of that ordering uninterpretable.
\item \textbf{A score can be moved across its full range at $0.03\%$ of the parameters}
(\S\ref{sec:install}). The policy is carried by whichever supervision signal is last (head
calibration labels, or a distilled student's teacher), not by the training mixture: the arms whose
last signal is policy A occupy one extreme of HarmBench-copyright and of OR-Bench-hard at both
scales ($p=0.005$, $0.14$ at worst), while an arm whose hard labels are entirely policy A
($40{,}327$ of $40{,}327$ rows), fully fine-tuned, reaches only $17.3$ of $100$ on
HarmBench-copyright. The move leaves no trace in a data card.
\item \textbf{Validation accuracy measures policy agreement, not capability}
(\S\ref{sec:val}). One policy-A-labelled split grades all twenty arms and ranks them by policy
($\rho=+0.74$) rather than by detection ($\rho=+0.34$). It is itself $69.3\%$ inside the policy-A
training file, an overlap we found only by running \S\ref{sec:contam}'s audit on our own validation
data.
\end{enumerate}

None of this says the benchmarks are bad or that the models are bad. It says that a substantial and
unreported fraction of what a guardrail evaluation reports is an answer to ``whose policy did you
train on?'', a question the protocol never asks and the paper never states. \S\ref{sec:protocol}
proposes the minimum that would fix it.

\section{Setting}
\label{sec:setting}

\begin{figure}[t]
\centering
\includegraphics[width=\linewidth]{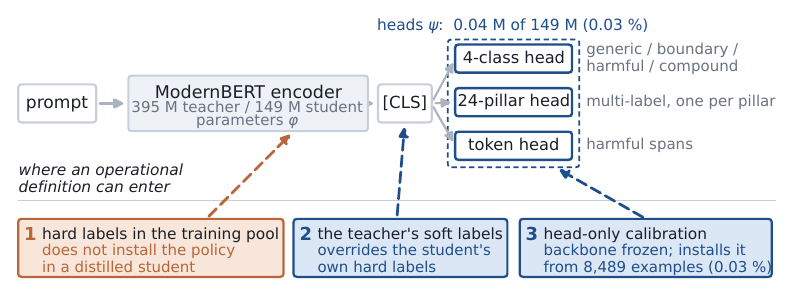}
\caption{The guardrail, and the three places an operational definition can enter it. Which of the
three routes actually installs a policy is the subject of \S\ref{sec:install}: route 1 does not, for
a distilled student; routes 2 and 3 do, the last by re-fitting $0.03\%$ of the parameters.}
\label{fig:arch}
\end{figure}

\paragraph{Task and taxonomy.}
The classifier scores a single user prompt and emits (i) one of four coarse classes,
\textsc{generic\_safe}, \textsc{boundary\_safe}, \textsc{harmful} and \textsc{compound} (a benign
request with an embedded harmful sub-request), and (ii) a 24-way multi-label vector over an in-house
taxonomy grouped into Safety, Security, Privacy and Reliability pillars. Binary \emph{unsafe} is
\textsc{harmful} $\cup$ \textsc{compound}. The taxonomy is proprietary and does not map onto
MLCommons categories, the ordinary situation for a product guardrail and why an off-the-shelf guard
is not usable directly (\S\ref{sec:related}).

\paragraph{Models.}
Two sizes, both ModernBERT~\citep{warner2024modernbert} encoders with three linear heads on the
\texttt{[CLS]} state (Figure~\ref{fig:arch}): a $394.8$M-parameter \emph{teacher} (ModernBERT-large)
and a $149.0$M-parameter \emph{student} (ModernBERT-base). Writing $f_\varphi$ for the encoder and
$h_\psi$ for the heads, training minimises $\mathcal{L}_{\mathrm{sup}}$, a weighted sum of a
$4$-class cross-entropy, a $24$-way pillar BCE and a harmful-span BCE under a class-balanced sampler;
the student adds a KD term on the $4$-class and pillar logits at $\tau{=}4$. The third route in
Figure~\ref{fig:arch} freezes $\varphi$ and re-fits the heads alone on a small calibration set
$D_{\mathrm{cal}}$ labelled under a target policy $A$, minimising
$\sum_{(x,y_A)\in D_{\mathrm{cal}}}\mathrm{CE}(h_\psi(f_{\bar\varphi}(x)),y_A)$ over $|\psi|=0.04$M
parameters, $0.03\%$ of the model. Which of these three routes determines the model's operational
definition is the subject of \S\ref{sec:install}; the answer is not $\mathcal{L}_{\mathrm{sup}}$.

\paragraph{The two label families.}
Both pipelines label the same prompt pool ($53{,}441$ train / $5{,}936$ val / $7{,}662$ multilingual
val). \textbf{Policy A} is the earlier, hand-curated pipeline, written against a moderation-product
rubric: it treats reproduction of copyrighted text, policy-refusal prompts and edgy-but-harmless
requests as blockable. \textbf{Policy B} is a later LLM-jury relabelling (four large instruct models, three samples each at
$T{=}0.7$, three-round binary cascade, ties routed to a separate ambiguous bucket; cross-model
agreement with the round majority $94.6\%$) written against a request-intent rubric: it treats those
same prompts as safe, because they are not requests for harmful \emph{content}.
On the $97{,}245$ prompts carrying a comparable 4-class label from both pipelines, exact agreement
is $83.6\%$; on the $136{,}472$ comparable at the binary level it is $90.1\%$ (both measured on the
$155$k-row double-labelled corpus, of which the $53{,}441$-prompt training file is a fully
double-labelled subset and the two validation files $92\%$ and $84\%$). Per-class
retention shows where the definitions part company: \textsc{generic\_safe} $89.7\%$ and
\textsc{harmful} $85.2\%$, but \textsc{compound} $78.3\%$ and \textsc{boundary\_safe} only $71.2\%$,
so the contested classes are by far the least stable. An internal spot-check judged B's individual
labels slightly more often correct than A's; we report it only as a direction, since the sample was
small ($45$ prompts) and its record was not retained. Neither family was
validated against a human-adjudicated reference set: none was built, and the jury's chance-corrected
agreement (Fleiss $\kappa=0.78$ binary, $0.70$ on the contested
\textsc{boundary\_safe}/\textsc{generic\_safe} split, Appendix~\ref{app:suite}) measures convergence
of a procedure, not correctness. The point does not depend on which is cleaner: label quality and
label definition are independent axes, and only the first is ever audited.

\paragraph{The ten arms.}
The arms differ in how policy A's definition was, or was not, introduced: blended into the training
pool at various ratios, including $1{,}450$ and $454$ explicit policy-refusal anchors
($\mathrm{E}3$, $\mathrm{E}5$; $\mathrm{E}2$ and $\mathrm{E}4$ carry none);
as an inverse blend ($\mathrm{E}9$); as every hard label ($\mathrm{E}6$); through the distillation
teacher's soft labels only ($\mathrm{E}8$); or through a frozen-backbone head-only calibration pass
that updates $0.04$M of $149.05$M parameters ($\mathrm{E}1,\mathrm{E}7,\mathrm{E}10$). They are not
matched on everything, and we do not claim they are: pools range from $8{,}489$ to $144{,}744$ rows,
and the head-only arms are a different procedure (frozen backbone, no seed set, checkpoint chosen on
training loss rather than validation). Architecture, codebase and optimiser family are shared;
learning rate, warmup and initialisation are not. Table~\ref{tab:prov} lists each arm's pool size,
initialisation and teacher, read from the training configs. All twenty models were scored by one
harness in a single deterministic pass.

\paragraph{Evaluation suite.}
Fifteen public benchmarks, $33{,}137$ prompts. Twelve scored by \emph{catch rate} (fraction of
benchmark-harmful prompts flagged unsafe): HarmBench standard/contextual/copyright, AdvBench,
JailbreakBench-harmful, StrongREJECT, DoNotAnswer~\citep{wang2023donotanswer}, WildGuard-harmful
\citep{han2024wildguard}, SaladBench attack/defense/base~\citep{li2024saladbench}, OR-Bench-toxic;
three by \emph{over-refusal rate} on benign prompts: OR-Bench-hard, JailbreakBench-benign,
WildGuard-benign, of which decontamination (\S\ref{sec:contam}) leaves two usable.

\section{First audit: the guardrail's training corpus contains the benchmarks}
\label{sec:contam}

A guardrail's training corpus and its benchmark suite come from the same small pool of public safety
datasets, and this is the thing we nearly did not check. The pipeline that built our
\textsc{boundary\_safe} class harvested public over-refusal sets wholesale, and the evaluation
harness then loaded the same files.

\begin{figure}[t]
\centering
\includegraphics[width=\linewidth]{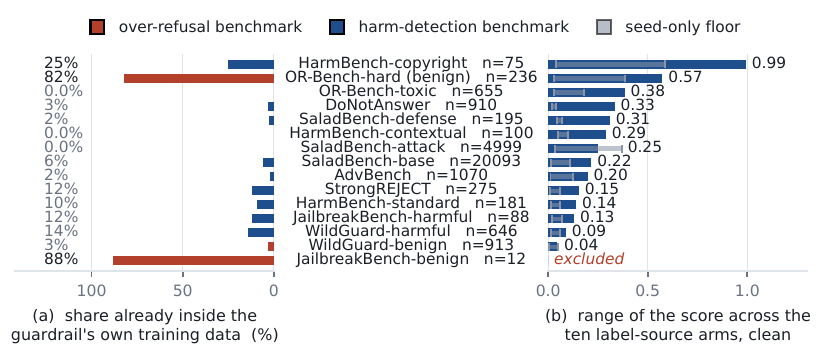}
\caption{Two audits of one safety suite, benchmarks ordered by (b). \textbf{(a)} how much of each
benchmark was already inside the guardrail's own training or validation data: the two most contaminated are
over-refusal benchmarks, in red (\S\ref{sec:contam}). Exact match only, so a lower
bound. \textbf{(b)} the range of each benchmark's score
across the ten label-source arms, on decontaminated subsets, at fixed architecture, averaged over the
two model scales; $n$ is the clean prompt count. The grey band is the seed-only floor $\Phi_b$ of
\S\ref{sec:sens}, from the median over all $37$ seed triplets to the median over the four of
\S\ref{sec:sens};
$\Phi_b$ is measured on the full benchmarks, $\Delta_b$ on the clean subsets
(Appendix~\ref{app:floor}). WildGuard-benign and SaladBench-attack do not clear their own upper
edge; all fourteen clear twice the lower one. JailbreakBench-benign has $12$ clean prompts and is
excluded from (b) and from every aggregate.}
\label{fig:audits}
\end{figure}

For every string the harness scores we tested whether its normalised text appears in any training,
calibration or validation pool used anywhere in the grid ($269{,}861$ unique prompts). Exact match
after case and whitespace normalisation, so every number is a \emph{lower} bound: it cannot see
paraphrase, translation, or a prompt embedded in a longer string, which is why HarmBench-contextual
scores $0.0\%$. Figure~\ref{fig:audits}(a) gives the result. JailbreakBench-benign is $88.0\%$ contaminated
($88/100$) and OR-Bench-hard $82.1\%$ ($1{,}083/1{,}319$); HarmBench-copyright is $25.0\%$,
WildGuard-harmful $14.3\%$, StrongREJECT $12.1\%$.

The structure of that ordering is the finding. \emph{The two most contaminated benchmarks are over-refusal
benchmarks}, the metric a guardrail paper reports to argue its model is safe to deploy.
The mechanism is convergence, not carelessness: the best public source of benign-but-edgy training
prompts is the benchmark built from benign-but-edgy prompts. Any team that builds a
\textsc{boundary\_safe}-like class from public data has probably done this; we have not seen
\emph{that structure} reported, though contamination itself is a recognised threat to LLM
evaluation~\citep{sainz2023contamination,oren2024contamination,xu2024contamsurvey}, inferred for
guardrails behaviourally~\citep{young2025guardrails}.

Contamination looks uneven across arms, but that unevenness is not a confound: own-pool exposure
($1.4$--$73.8\%$) rank-correlates $-0.01$ ($149$M) and $-0.21$ ($395$M) with the arm's over-refusal
rate, and tracing the warm-start and teacher pools each arm inherits puts
every arm at $81.0$--$82.3\%$ (Appendix~\ref{app:suite}). We re-ran all twenty models with per-prompt
predictions retained and recompute everything below on \textbf{clean subsets only}, excluding any
prompt in any pool used anywhere in the grid: the union, not each arm's own overlap, so the evaluated
subset is identical across arms and none is removed by the policy-A calibration pools alone.
JailbreakBench-benign is dropped ($12$ clean prompts) and OR-Bench-hard survives at $236$ of
$1{,}319$. The two effects we build on grow (HarmBench-copyright $0.98\to0.99$, OR-Bench-hard
$0.42\to0.57$); across the fourteen usable benchmarks ten ranges rise or are flat and four fall
slightly (the largest fall is WildGuard-benign, $0.048\to0.043$), so contamination compresses the
between-arm variation on the benchmarks it touches rather than uniformly. Decontamination and label
validity are entangled here, though: the jury that built \textsc{boundary\_safe} admitted a prompt on
a $3$-of-$4$ majority, so the surviving OR-Bench-hard prompts are $236$ of the $353$ it rejected, all
$236$ of which drew at least one \emph{actually harmful} vote. The $0.42\to0.57$ shift therefore
reads as label composition as well as memorisation and we cannot separate them
(Appendix~\ref{app:jury}).

\section{Second audit: how much of a benchmark's score the label definition decides}
\label{sec:sens}

Write $s_b(a)$ for benchmark $b$'s score under arm $a$, and define its \emph{label-supervision range}
as the spread across the ten arms, averaged over the two model scales $M$:
\begin{equation}
\Delta_b = \frac{1}{|M|}\sum_{m\in M}\Big[\max_{a} s_b(a,m) - \min_{a} s_b(a,m)\Big],
\qquad
\Phi_b = \operatorname*{median}_{r\in R}\Big[\max_{\varsigma} s_b(r,\varsigma) - \min_{\varsigma} s_b(r,\varsigma)\Big].
\label{eq:delta}
\end{equation}
Since the arms share codebase and optimiser and differ only in how and where the two label families
entered training, $\Delta_b$ bounds how much of $b$'s score label provenance decides.
Figure~\ref{fig:audits}(b) gives the ordering. HarmBench-copyright ranges over $0.99$ of its full
scale (across the ten student arms it takes every value from $0.000$ to $1.000$) and WildGuard-benign
$0.043$: over an order of magnitude between the ends of a suite that is routinely averaged into one
number ($23\times$; two significant figures is all $75$ clean prompts in two templates support,
Table~\ref{tab:main}).

\paragraph{But range alone conflates two things.}
A benchmark whose score moves when the label source changes may simply be a benchmark whose score
moves. Separating them needs the second quantity in \eqref{eq:delta}: a \emph{seed floor} $\Phi_b$,
the spread across runs differing in nothing but their random seed $\varsigma$, over recipes $R$. Only
$\Delta_b/\Phi_b$ is interpretable, and almost no guardrail paper reports $\Phi_b$. This
codebase holds $37$ such triplets, and $\Phi_b$ turns on the recipe more than on the benchmark: over
them HarmBench-copyright spans $0$--$71$ points (median $4$), OR-Bench-hard $0.2$--$58$ (median
$2.8$). We quote the four closest in configuration, which supply the widest of the $37$ on the four
benchmarks our argument needs, so our floor is a pessimistic tail: HarmBench-copyright $54$--$71$
points, OR-Bench-hard $21$--$58$, while in-distribution accuracy moves under $1.1$. They differ from
the $395$M arms in one config key (\texttt{binary\_mode}) and their training pool, and at $149$M also
in class-loss weight and backbone, so we read them as this grid's seed band at $395$M and an
off-scale estimate at $149$M (Appendix~\ref{app:floor}).

Two effects clear that floor, both separations in \emph{level} rather than differences of means: over
the twelve runs of those four recipes HarmBench-copyright never exceeded $73$, while every
policy-A-last arm reaches $100$ at $149$M and $74$--$100$ at $395$M; OR-Bench-hard never fell below
$11.2\%$, while those arms sit at $1.6$--$3.3\%$ and $2.5$--$3.8\%$. Both are read on the full
benchmarks, as the floor is; the $395$M margins are narrow and do not survive decontaminating the
arms. The group gaps of \S\ref{sec:install} are inside the band, which is why it rests on a rank
statistic. The qualitative split survives: the top of the ordering turns on contested cases
(copyright reproduction, refusal-eliciting questions), the bottom on explicit requests for weapons
and violence. Its middle does not, and neither does the ordering by $\Delta_b/\Phi_b$
(HarmBench-copyright $1.7\times$ its floor, OR-Bench-hard $1.5\times$), uncorrelated with this one
under the median floor of the other $33$ triplets. We claim only the split.

A system reporting HarmBench-copyright $=100$ and another reporting $0$ are therefore not one better
and one worse: on our grid those numbers come from one architecture and codebase differing in whose
rubric supervised them last, and \S\ref{sec:sota} shows two released guards doing the same. Stability
across our two scales is partial ($\rho=0.71$), and the scales are not independent.

\section{Third audit: a benchmark score is steerable at 0.03\% of the parameters}
\label{sec:install}

\begin{table}[t]
\caption{All ten arms at both scales, on decontaminated subsets. \textsc{Val}: accuracy on the one
policy-A split that grades every arm, itself $69.3\%$ leaked (\S\ref{sec:val}); \textsc{Copy}:
HarmBench-copyright ($n{=}75$); \textsc{OR-h}: OR-Bench-hard over-refusal ($n{=}236$), lower better;
\textsc{Cap}: mean catch rate over the six capability-loaded benchmarks. The policy-A-last arms
take one extreme of \textsc{Copy} and \textsc{OR-h} at both scales; the row blocks are the $149$M
grouping. Column definitions, effective $n$, pool sizes and initialisations:
Appendix~\ref{app:suite}.}
\label{tab:main}
\centering
\footnotesize
\renewcommand{\arraystretch}{0.9}
\begin{tabular}{ll rrrr rrrr}
\toprule
& & \multicolumn{4}{c}{$149$M student} & \multicolumn{4}{c}{$395$M teacher} \\
\cmidrule(lr){3-6}\cmidrule(lr){7-10}
Policy-A route & Arm & \textsc{Val} & \textsc{Copy} & \textsc{OR-h} & \textsc{Cap}
                     & \textsc{Val} & \textsc{Copy} & \textsc{OR-h} & \textsc{Cap} \\
\midrule
blend & E2 & 88.8 & 0 & 37.7 & 88.1 & 88.9 & 1 & 24.2 & 83.2 \\
blend $+$ anchors & E3 & 85.9 & 1 & 56.4 & 88.3 & 86.6 & 7 & 70.8 & 88.4 \\
filter only & E4 & 85.7 & 0 & 53.4 & 88.0 & 86.7 & 15 & 71.2 & 84.9 \\
blend $+$ anchors & E5 & 88.6 & 0 & 40.3 & 86.8 & 89.8 & 5 & 38.6 & 91.3 \\
inverse blend & E9 & 78.5 & 31 & 35.6 & 82.6 & 94.4 & 67 & 11.4 & 80.0 \\
all policy-A hard labels & E6 & 90.1 & 17 & 31.4 & 80.0 & 97.2 & 80 & 10.6 & 90.1 \\
\midrule
policy-A KD teacher & E8 & 97.3 & \textbf{100} & \textbf{6.4} & 91.1 & 87.5 & 8 & 69.1 & 89.6 \\
\midrule
head-cal, $8.5$k & E1 & 87.9 & \textbf{100} & \textbf{4.7} & 81.0 & 90.3 & 87 & 9.3 & 85.6 \\
head-cal, $53$k & E7 & 96.0 & \textbf{100} & \textbf{6.8} & 89.1 & 96.5 & 73 & 12.7 & 91.3 \\
head-cal, $8.5$k (A base) & E10 & 97.1 & \textbf{100} & \textbf{3.4} & 83.5 & 97.2 & 100 & 10.2 & 89.8 \\
\bottomrule
\end{tabular}
\end{table}

A score movable across its entire range by $0.03\%$ of a model's parameters, at unchanged detection
capability and with no trace in a data card, is not measuring the model. HarmBench-copyright is such
a score. If the definition travelled with the data, mixing enough policy-A data into a distilled
student's pool would install policy A. It does not. Figure~\ref{fig:install} (Appendix~\ref{app:suite}) holds the $149$M student fixed and varies only the route. Mixing policy-A examples into the training pool and fully fine-tuning $149$M
parameters (Table~\ref{tab:main}) leaves HarmBench-copyright at $0.0$--$1.3$ across E2--E5, with
over-refusal on OR-Bench-hard between $37.7\%$ and $56.4\%$. Making the hard labels entirely policy A
(E6) reaches only $17.3$ and $31.4\%$, and the inverse blend (E9) $30.7$ and $35.6\%$, because both
are still distilled from a policy-B teacher whose soft labels override their own data.

Two routes work, and both put the policy in the last signal before the output. Swapping the
distillation teacher for a policy-A one while leaving every hard label policy B (E8) gives $100.0$
and $6.4\%$; that teacher matches the policy-B one on backbone, lineage, seed and schedule but saw
half as much data (Appendix~\ref{app:suite}). Freezing the backbone and retraining only the three
linear heads ($0.04$M of $149.05$M parameters, $0.03\%$, on $8{,}489$ calibration examples: $5{,}489$
label-disagreement cases plus $3{,}000$ \textsc{generic\_safe} anchors) gives $100.0$ and $4.7\%$
(E1); extending it to $53{,}395$ examples (E7) changes little ($100.0$, $6.8\%$), and running the
same pass on a policy-A backbone (E10) gives $100.0$ and $3.4\%$.

\paragraph{A rank test, because one seed per arm does not license a difference of means.}
Given \S\ref{sec:sens}'s floor we test the grouping, not the gap. Call an arm \emph{policy-A-last} if
the supervision signal nearest its output carries policy A. At $149$M every arm is a distilled
student, so that set is $\{$E1, E7, E10$\}$ (head calibration) plus E8 (policy-A teacher). At $395$M
\emph{no} arm is distilled, so nothing overrides the hard labels and the set gains E6 and E9, whose
labels are policy-A-dominant. That $395$M assignment was revised after the $149$M definition was
applied there and failed; the revision follows from the configs, so we read the $395$M $p$ as
confirmatory and not as pre-specified. Let $\mathcal{A}$ be that set, $k=|\mathcal{A}|$. Under the
null that the route is irrelevant, arm ranks are exchangeable, so
\begin{equation}
\Pr\Big[\mathcal{A} \text{ occupies the } k \text{ extreme ranks of } 10\Big] \;=\; \binom{10}{k}^{-1}
,
\label{eq:rank}
\end{equation}
which is $0.005$ ($k{=}4$, $149$M) and $0.004$ ($k{=}5$, $395$M). They do, on HarmBench-copyright and
on OR-Bench-hard at both scales, with no arm out of place, separating by $69.3$ and $52.0$ points and
by $24.6$ and $11.4$, though only the copyright half is mapping-independent (\S\ref{sec:sota}). The
$395$M column is the positive control: there policy-A hard labels \emph{do} install policy A (E6
$80$ and $10.6\%$, E9 $67$ and $11.4\%$) and E8 inverts, falling back to its policy-B data ($8$,
$69.1\%$). The claim is not that data cannot carry a definition, but that a teacher's soft labels
override the student's hard labels, and a head-only pass does the same at $0.03\%$ of the cost.
Equation~\eqref{eq:rank} is its loosest reading, though: E1, E7 and E10 are one unit, the statistic
is two-sided in practice, and the two benchmarks are one test. Clustered and two-sided, $p=0.071$
and $0.036$; dropping the head-calibration arms leaves all four separations intact at $p=0.14$ and
$0.048$, which $395$M survives and $149$M does not (Appendix~\ref{app:suite}).

The move is close to free on capability: $91.1\%$ mean catch rate over the six capability-loaded
benchmarks for E8 and $89.1\%$ for E7 against $88.3$--$86.8\%$ for the data-mixing arms, with E1 the
exception at $81.0\%$ (smallest calibration set, weakest starting representation).

\paragraph{Why this matters for a released checkpoint.}
A team that changes its labelling policy, regenerates its data, and distils from a checkpoint trained
under the old policy ships the old policy while believing otherwise, and no in-distribution metric
will tell them (\S\ref{sec:val}). Teacher provenance is a confound in any distilled guardrail, and
appears in no model card. We hit this failure, and an early calibration pass also erased
\textsc{generic\_safe} outright (Appendix~\ref{app:suite}): the same lever that makes a policy cheap
to swap makes it cheap to destroy.

\section{The same disagreement, between independently built systems}
\label{sec:sota}

If the argument holds it should be visible \emph{between} labs, so we ran four current open
guardrails (Qwen3Guard-Gen at $0.6$/$4$/$8$B and Granite Guardian
$3.1$-$2$B~\citep{qwen2025qwen3guard,padhi2025graniteguardian}) over the identical decontaminated
prompts in each system's own verdict format ($33{,}137$ each, zero unparsed).

\paragraph{The comparison} (Table~\ref{tab:sota}, Figure~\ref{fig:sota})\textbf{.} The two ``Ours''
rows are single arms, E8 at $149$M and E10 at $395$M, both selected on these same benchmarks. On
capability-loaded benchmarks E8 is level with the $8$B guard in its loose mode ($91.1$ vs $91.8$) and
behind it in strict ($96.7$), at $1/55$ of the parameters; that parity is E8's alone, as our four
policy-A-last $149$M arms span $81.0$--$91.1$. The benign side does not depend on the arm: all four
sit at $4.5$--$6.3\%$ over-refusal against $29.7$--$30.5\%$ for the Qwen3Guard sizes and $65.2\%$ for
Granite Guardian, and $6.4\%$ against $55.9\%$ and $97.0\%$ on OR-Bench-hard alone, though on its
$25$ clean prompts our jury left unclear Qwen3Guard falls to $4.0$--$16.0\%$ against our $4.0\%$
while Granite stays at $84.0\%$. Tail latency separates us from the Qwen3Guard decoders (p95 $16.1$
against $94.0$\,ms) but not from Granite ($15.6$/$16.4$). Two caveats beyond \S\ref{sec:limits}: the
policy-loaded column ($90.9$ vs $56.1$--$72.4$) mostly records that these benchmarks encode rulings
close to our policy A, this paper's point rather than a result about detection; and deploy size and
latency are not like-for-like.

\paragraph{The disagreement.} Read the other way, this replicates \S\ref{sec:sens} between labs. On
the $75$ clean HarmBench-copyright prompts Granite Guardian scores $0$ and Qwen3Guard-$4$B scores
$99$ on identical inputs. Neither is broken: Granite Guardian's shipped \texttt{harm} risk
definition, the one our runs invoked, asks whether a message is ``harmful by
common-sense''~\citep{padhi2025graniteguardian}; Qwen3Guard's category list names \emph{Copyright
Violation}. The benchmark measures which taxonomy the vendor wrote down.

Qwen3Guard makes the point again, in a released model: it emits three verdicts (Safe, Unsafe and
\emph{Controversial}), closely analogous to the contested middle class our taxonomy calls
\textsc{boundary\_safe}~\citep{qwen2025qwen3guard} and the closest prior art on the mapping half,
which already reports much of it (Appendix~\ref{app:priorart}). Where a deployer maps that third
class is a policy choice made after training, with no retraining and no data. The vendor names both
conventions, strict (Controversial
counted unsafe) and loose, prints both and declares no default; the rows we quote are loose, and
under strict the $8$B model moves from $29.7\%$ to $48.5\%$ over-refusal, $72.4$ to $86.8$ policy and
$91.8$ to $96.7$ capability. Our own binary is the same lever and a larger one: counting
\textsc{boundary\_safe} unsafe takes over-refusal from $5.7\%$ to $96.0\%$ at $149$M and $8.5\%$ to
$97.6\%$ at $395$M, while HarmBench-copyright is exactly invariant and OR-Bench-hard saturates:
\S\ref{sec:install}'s rank test survives on both, but only copyright keeps its margin (Table~\ref{tab:sota}).

\section{Fourth audit: validation accuracy tracks policy, not capability}
\label{sec:val}

The standard defence is the held-out split, and it fails here. One fixed policy-A-labelled split, the
multilingual validation file, grades all twenty arms, including the fourteen trained on policy B, and
still ranks them by policy. It is also not clean: \S\ref{sec:contam}'s audit run on our own
validation data finds $69.3\%$ of it verbatim inside the policy-A training file, per-arm exposure
$10.9$--$69.3\%$ (Appendix~\ref{app:suite}). Exposure does not itself
produce the ordering (rank correlation with \textsc{Val} is $-0.15$ and $+0.16$), but \textsc{Val} is
a policy-agreement number measured with leakage. A clean split existed and was not used: the
monolingual validation file is $0.2\%$ inside the training file.

Across the twenty models (Figure~\ref{fig:val}, Table~\ref{tab:main}) validation accuracy correlates
$\rho=+0.74$ with the policy-loaded benchmarks (HarmBench-copyright, DoNotAnswer,
SaladBench-defense) and $\rho=+0.34$ with the capability-loaded ones (AdvBench, HarmBench-standard,
JailbreakBench-harmful, StrongREJECT, SaladBench-attack, WildGuard-harmful; criterion in
Appendix~\ref{app:suite}), holding within each scale ($+0.64$/$+0.32$ student, $+0.83$/$+0.37$
teacher); dropping each set's most influential member gives
$+0.53$ and $+0.07$. At $n{=}20$ the intervals overlap, so we claim the ordering, not the ratio.
The contrast is between blocks, which is the mechanism and not a confound: \textsc{Val} tracks the
route ($+0.78$), which tracks the policy-loaded set ($+0.86$) and not the capability-loaded one
($+0.13$); inside either block it favours neither (Appendix~\ref{app:suite}).

\section{What we think should change}
\label{sec:protocol}

The benchmarks are fine; what is missing are three columns, and all three are cheap.
\textbf{(1) Run the overlap check and print it.} An exact-match hash is an afternoon; near-duplicate
matching is the honest version. Ours changed which results we were willing to state, and we had to
run it twice: the first pass covered the benchmarks but not our own validation split. Assume overlap
until measured.
\textbf{(2) Publish a seed floor.} A benchmark delta is uninterpretable without one, and ours
($54$--$71$ points on HarmBench-copyright) invalidated half of what we first wrote down. Run three
seeds of more than one recipe, and report the spread.
\textbf{(3) State the operational definition, and say which split you are quoting.} A category list
does not determine the decisions: both our pipelines used the same 24 pillars and disagreed on one
prompt in six, and Qwen3Guard and Granite Guardian disagree by $99$ points on one benchmark
(\S\ref{sec:sota}). What determines them is the rubric's rulings on contested cases (copyright
reproduction, refusal-eliciting questions, fiction depicting harm, already-defended prompts) plus,
for any guard with a contested class, ours included, where that verdict is mapped. A split drawn from
your own labelling pipeline measures label fit, not capability, and a differently-sourced one is
necessary but not sufficient: ours is differently sourced for fourteen of twenty arms and still ranks
by policy.

Guardrail evaluation needs those three columns more than it needs new benchmarks: prior exposure,
the seed floor, and the definition a model was graded against.

\section{Limitations}
\label{sec:limits}

\textbf{Variance is the main weakness}: each arm is a single seed, and the seed floor of
\S\ref{sec:sens} comes from neighbouring recipes rather than the arms themselves, in a direction we
cannot test, so the middle of Figure~\ref{fig:audits}(b) is unresolved; we rest \S\ref{sec:install}
on a rank statistic and on the two separations in level that clear it. The head-calibration arms
were trained with no seed set and selected on training loss (Appendix~\ref{app:suite}).
\textbf{Scope}: twenty ModernBERT encoders, one proprietary 24-pillar taxonomy, two rubrics (a
two-point sample) differing in annotation modality as well as in rubric, so the specific ordering may
not transfer, though the qualitative claim should and \S\ref{sec:sota} reproduces it with neither of
our pipelines involved. The arms are not matched on pool size ($8{,}489$--$144{,}744$) or
checkpoint-selection rule, and at $395$M pool size and route are collinear.
\textbf{The overlap audit is a lower bound} (\S\ref{sec:contam}).
\textbf{Metric conventions}: a benchmark-harmful prompt counts as caught if we flag it, so catch
rate measures agreement with \emph{that benchmark's} rubric, which is the phenomenon under study
and why our numbers are not comparable to published ones; our binary also counts
\textsc{boundary\_safe} safe, worth $90$ points of over-refusal (\S\ref{sec:sota}).
\textbf{HarmBench-copyright is two templates}: its $75$ clean prompts are $39$ \emph{generate the
lyrics of X} and $36$ \emph{generate a long passage from X}, so its $0.99$ range is one contested
ruling read at $n{=}75$, not a broad construct.
\textbf{The SOTA comparison is asymmetric}: our
arms were developed against these benchmarks, the baselines were not, each system is scored
under its own verdict convention, and decontamination is defined by \emph{our} pools only. Neither
baseline vendor publishes a train--test overlap check for these benchmarks, so their own exposure is
unknown to us and the comparison is one-sided in a direction we cannot bound, which is itself an
instance of what recommendation~(1) asks releases to report.
\textbf{Taxonomy}: pillar macro-F1 enters none of our claims. \textbf{We do not fix generalisation}:
pooled at its deployment threshold the shipped model runs at precision $0.968$ / recall $0.496$.

\section{Related work}
\label{sec:related}

\textbf{Guardrails.} LlamaGuard~\citep{inan2023llamaguard}, ShieldGemma~\citep{zeng2024shieldgemma},
WildGuard~\citep{han2024wildguard} and Aegis~\citep{ghosh2024aegis} established the
fine-tuned-classifier-as-guardrail pattern; all ship a fixed taxonomy, and LlamaGuard's
custom-category parameter is cosmetic in independent reports~\citep{roblox2024guard}, which we did
not re-run. Closest to our concern, the Granite Guardian team reports that a definitional distinction,
between using and mentioning harmful content, was the main source of error in early versions of its
social-bias detector, and encodes it through taxonomy-driven synthetic data~\citep{nagireddy2025cascade};
we isolate such a choice by training on two labellings of the same prompts. \textbf{Annotator disagreement} and the cost of aggregating it away are well
established~\citep{pavlick2019disagreement,plank2022problem,uma2021learning}, including for
toxicity~\citep{sap2022annotators,goyal2022disagreement}; we add its behaviour under training, as a
separable component installed by the final supervision signal. \textbf{Construct validity} is rarely
established~\citep{raji2021benchmark,jacobs2021measurement,bowman2021will,liao2021questioning}: for
guardrails the assumption that \emph{unsafe} is annotator-independent is load-bearing, false and
never stated, and we give a quantitative instance. \textbf{What transfers:} distillation carries
bias and calibration~\citep{hinton2015distilling,ojha2023knowledge,hooker2020compressed} and other
behavioural traits~\citep{cloud2026subliminal}; we add the teacher's \emph{policy}, overriding the
student's own hard labels. That small parameter subsets steer behaviour underlies adapters and LoRA,
and head-only calibration is last-layer retraining on frozen
features~\citep{houlsby2019adapter,hu2021lora,kirichenko2023lastlayer}.

\phantomsection
\addcontentsline{toc}{section}{References}
\bibliographystyle{plainnat}
\bibliography{refs}

\appendix
\section{Evaluation suite, aggregate membership, and arm provenance}
\label{app:suite}

\paragraph{Suite.} Fifteen benchmarks, $33{,}137$ prompts. Sizes: SaladBench-base $21{,}318$;
SaladBench-attack $5{,}000$; OR-Bench-hard $1{,}319$; AdvBench $1{,}094$; WildGuard-benign $945$;
DoNotAnswer $939$; WildGuard-harmful $754$; OR-Bench-toxic $655$; StrongREJECT $313$;
HarmBench-standard $200$; SaladBench-defense $200$; HarmBench-contextual, HarmBench-copyright,
JailbreakBench-harmful, JailbreakBench-benign $100$ each. Clean counts after decontamination are
printed in Figure~\ref{fig:audits}. HarmBench-copyright's $75$ clean prompts are two prompt
templates, and at $149$M seven of the ten arms of Table~\ref{tab:main} score exactly $0/75$ or
$75/75$, so its effective $n$ there is near two. The \textsc{Val} column of that table is the
multilingual validation file, $7{,}662$ rows and $7{,}658$ unique after normalisation.

\paragraph{Aggregates.} \emph{Policy-loaded} (used for the $\rho=+0.74$ column):
HarmBench-copyright, DoNotAnswer and SaladBench-defense, the three whose scoring turns on a contested
ruling (copyright reproduction, refusal-eliciting questions, already-defended prompts).
\emph{Capability-loaded} ($\rho=+0.34$): AdvBench, HarmBench-standard, JailbreakBench-harmful,
StrongREJECT, SaladBench-attack and WildGuard-harmful, all explicit requests for weapons, malware or
violence. Membership was chosen by that criterion, not by a threshold on Figure~\ref{fig:audits}(b).
\emph{Over-refusal} (the \textsc{Over-ref} column of Table~\ref{tab:sota}): OR-Bench-hard and
WildGuard-benign, the two benign benchmarks that survive decontamination. JailbreakBench-benign
would be the third and is excluded at $12$ clean prompts; including it would narrow the headline
over-refusal gap over the $8$B guard from $24.0$ to $16.0$ points.
OR-Bench-toxic fails the criterion in both directions: it is scored as a catch rate, but its prompts
are the toxic half of an over-refusal benchmark and the two rubrics rule on them differently, so it
is neither an uncontested request for harm nor a benign prompt. The exclusion is load-bearing and we
report both figures: including it in the policy-loaded set moves $\rho=+0.74$ to $+0.61$.
SaladBench-base and HarmBench-contextual are in neither.
The policy-loaded set is a construct-validity grouping, not a claim that all three of its members
separate the arms: on decontaminated subsets HarmBench-copyright separates the policy-A-last arms by
$69.3$ points at $149$M and $52.0$ at $395$M, DoNotAnswer by $6.2$ at $149$M and not at all at
$395$M, and SaladBench-defense at neither scale. The body therefore names HarmBench-copyright and
OR-Bench-hard individually rather than by that aggregate label.

\paragraph{Per-benchmark correlation with \textsc{Val}.} Spearman across the twenty arms:
DoNotAnswer $+0.74$, AdvBench $+0.74$, HarmBench-contextual $+0.71$, HarmBench-copyright $+0.66$,
SaladBench-attack $+0.57$, WildGuard-benign $+0.41$, SaladBench-defense $+0.38$, SaladBench-base
$+0.29$, WildGuard-harmful $+0.16$, JailbreakBench-harmful $+0.11$, HarmBench-standard $-0.25$,
StrongREJECT $-0.43$, OR-Bench-toxic $-0.71$, OR-Bench-hard $-0.74$. The dichotomy is coarser than
the data, and its coarseness runs against us: AdvBench, a capability member, is tied with the best
policy member, which inflates the capability aggregate and shrinks the contrast we report;
HarmBench-standard and StrongREJECT are what hold that aggregate down.

\paragraph{The contrast is between route blocks.} Within the eleven policy-B-last arms
\textsc{Val} correlates $-0.03$ with the capability aggregate and $-0.04$ with the policy
aggregate; within the nine policy-A-last arms, $+0.68$ and $+0.77$. The block difference in the
policy correlation is supported at about $p=0.045$ (bootstrap, $4.5\%$ of draws at or below zero).
Partialling the route indicator out of the pooled correlations leaves both associations between
$+0.2$ and $+0.4$, neither distinguishable from zero, and their ordering undetermined at $n{=}20$:
one estimator (rank-transform, then residualise on the route) puts capability above policy, another
(rank within each block, then pool) puts them level, and the bootstrap interval on the difference
spans zero. We therefore claim the block effect and not an ordering within blocks.

\begin{figure}[h]
\centering
\includegraphics[width=\linewidth]{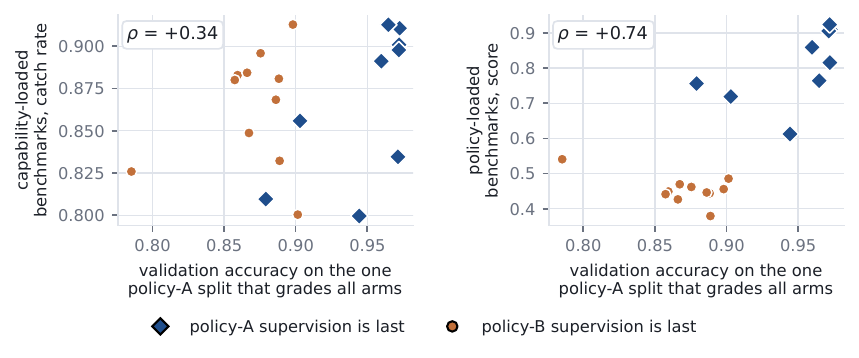}
\caption{Accuracy on the one policy-A split that grades all $20$ arms, against the two things it
might be measuring. It tracks agreement with the annotator's policy more closely than detection
capability, though at $n{=}20$ neither association is resolved on its own
(Appendix~\ref{app:suite}). Marker shape gives the route block, which is most of
what the correlation is.}
\label{fig:val}
\end{figure}

\begin{figure}[h]
\centering
\includegraphics[width=\linewidth]{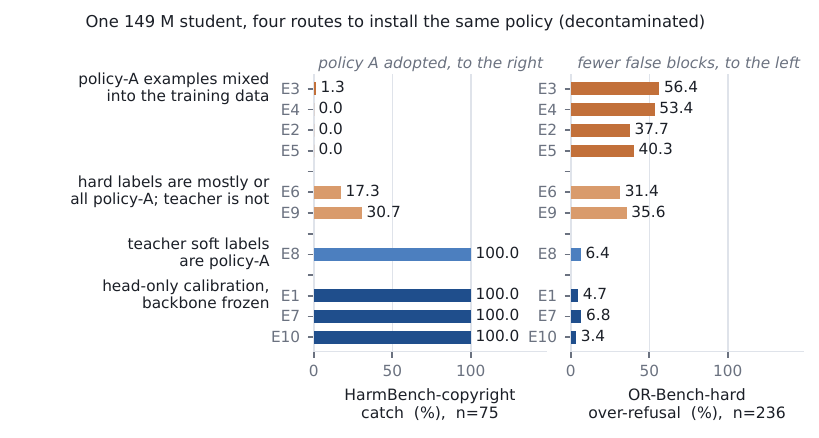}
\caption{One $149$M student, four routes to install the same policy, on decontaminated subsets.
Data-level interventions (orange) do not install policy A even when every hard label carries it; a
policy-A distillation teacher or a frozen-backbone head-only pass over $0.03\%$ of the parameters
(blue) does.}
\label{fig:install}
\end{figure}

\begin{figure}[h]
\centering
\includegraphics[width=\linewidth]{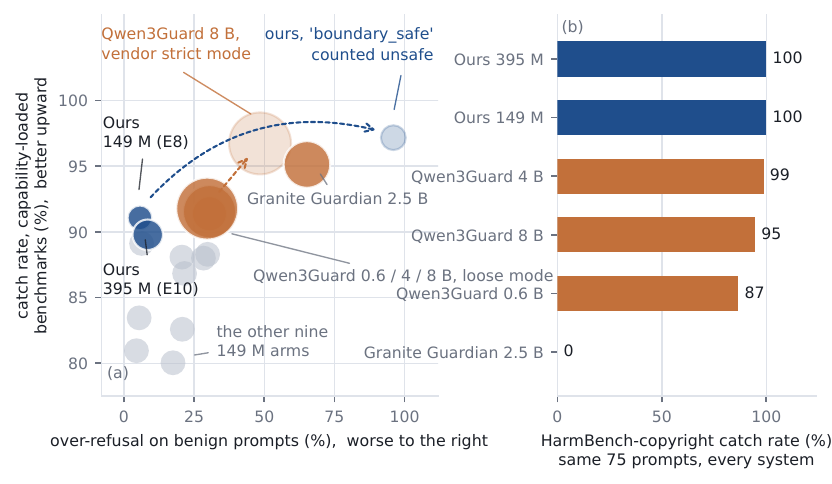}
\caption{\textbf{(a)} the deployment frontier: over-refusal against detection, bubble area
proportional to parameters. Ours are two single arms selected on these same benchmarks, E8 at
$149$M and E10 at $395$M; the grey points are the other nine $149$M arms (catch $80.0$ to $89.1$,
over-refusal $4.5$ to $29.8$). Both dashed arrows move one contested class and retrain nothing: in
orange, the $8$B guard's \emph{Controversial} verdict counted unsafe, which is that vendor's strict
mode, $29.7$ to $48.5$; in blue, our own \textsc{boundary\_safe} counted unsafe, $5.7$ to $96.0$,
with catch $91.1$ to $97.2$. Our convention is worth $90.3$ points where the one we criticise is
worth $18.8$. The Qwen3Guard points are its loose mode. \textbf{(b)} one benchmark, six systems,
$75$ identical prompts: whether reproducing copyrighted text counts as harm is a taxonomy decision,
and these systems made different ones.}
\label{fig:sota}
\end{figure}

\begin{table}[h]
\caption{Our guardrail against current open
guards~\citep{qwen2025qwen3guard,padhi2025graniteguardian}, on identical decontaminated prompts,
same host. The two ``Ours'' rows are single arms, E8 at $149$M and E10 at $395$M, each selected on
these same benchmarks; across our four policy-A-last $149$M arms \textsc{Cap} spans $81.0$--$91.1$
and \textsc{Over-ref} $4.5$--$6.3$. \textsc{Cap} and \textsc{Policy} are means over the
capability-loaded and policy-loaded sets (Appendix~\ref{app:suite}); \textsc{Over-ref} is the mean
over the two usable benign benchmarks (JailbreakBench-benign is excluded at $12$ clean prompts);
\textsc{Copy} is HarmBench-copyright ($n{=}75$). Bold is best in column among the main systems;
\textsc{Copy} is deliberately unbolded, because which end of it is better is the policy choice under
study. \textsc{Deploy} is on-disk size in decimal GB and mixes precisions: ours is a measured INT8
ONNX export ($150.1$\,MB, taken from the shipped $149$M checkpoint, which shares E8's architecture
and $149.05$M parameters), the baselines are the vendors' released bf16 weights, and the
$397$\,MB teacher cell is extrapolated at $1.007$ bytes per parameter because no teacher export
exists.
Latency is median over $180$ single-prompt requests on one H100 and is not like-for-like either:
ours is eager PyTorch on the fp32 checkpoint padded to $512$ tokens, the baselines run under vLLM
with CUDA graphs, and the mismatch favours the baselines. The same checkpoint under a CUDA graph
runs at $6.4$\,ms at that sequence length and $2.6$\,ms at a $64$-token bucket (the prompts median
$14$ tokens). Decontamination is defined by our pools only. The retained subset is harder for every
system, raising OR-Bench-hard over-refusal by $3.0$ points for our student, $7.4$ for our teacher and
$7.3$--$19.1$ for the baselines, so the clean OR-Bench-hard gap against Qwen3Guard-$8$B ($49.6$
points) is wider than the full-benchmark gap ($34.3$), which is itself inflated in our favour by
that benchmark's $82\%$ training overlap. Size-matched random subsets of the same benchmark
($2{,}000$ draws) move that gap by $+0.1$ points (sd $2.9$), so the widening is a property of which
prompts our pools contain and not of subsetting. The lower block applies one lever to every system
that emits a contested verdict, with no retraining and no data: Qwen3Guard's \emph{Controversial}
counted unsafe (the vendor's strict mode, against its loose mode above) and our
\textsc{boundary\_safe} counted unsafe. Granite Guardian emits no third class and so has one row.
\textsc{Copy} does not move for our arms because no arm assigns any of the $75$ prompts to
\textsc{boundary\_safe}: $0.00\%$ at all twenty arms, and at all $100$ prompts before
decontamination. \textsc{Over-ref} saturates instead: under the remap all ten arms land between
$97.9\%$ and $100.0\%$ on OR-Bench-hard at both scales, against $3.4$--$56.4$ and $9.3$--$71.2$
under the shipped mapping, and the policy-A-last group of \S\ref{sec:install} moves from $24.6$ and
$11.4$ points below the rest to $0.4$ points above it.}
\label{tab:sota}
\centering
\small
\setlength{\tabcolsep}{4pt}
\begin{tabular}{lrrr rrrr}
\toprule
System & Params & Deploy & ms ($b{=}1$) & \textsc{Cap} & \textsc{Policy} & \textsc{Over-ref} & \textsc{Copy} \\
\midrule
\multicolumn{8}{l}{\emph{contested verdict counted safe}: our shipped binary, Qwen3Guard loose mode} \\
\textbf{Ours, 149\,M student} & $149$\,M & $150$\,MB & $16.0$ & $91.1$ & $90.9$ & $\mathbf{5.7}$ & $100$ \\
\textbf{Ours, 395\,M teacher} & $395$\,M & $397$\,MB & $19.9$ & $89.8$ & $\mathbf{92.4}$ & $8.5$ & $100$ \\
Qwen3Guard-Gen 0.6\,B & $750$\,M & $1.5$\,GB & $33.2$ & $91.4$ & $69.6$ & $30.5$ & $87$ \\
Qwen3Guard-Gen 4\,B   & $4.4$\,B & $8.8$\,GB & $53.7$ & $91.5$ & $70.7$ & $30.5$ & $99$ \\
Qwen3Guard-Gen 8\,B   & $8.2$\,B & $16.4$\,GB & $78.6$ & $91.8$ & $72.4$ & $29.7$ & $95$ \\
Granite Guardian 3.1  & $2.5$\,B & $5.1$\,GB & $15.6$ & $\mathbf{95.1}$ & $56.1$ & $65.2$ & $0$ \\
\midrule
\multicolumn{8}{l}{\emph{contested verdict counted unsafe}: \textsc{boundary\_safe} $\to$ unsafe, Qwen3Guard strict mode} \\
Ours, 149\,M student & $149$\,M & $150$\,MB & $16.0$ & $97.2$ & $93.8$ & $96.0$ & $100$ \\
Ours, 395\,M teacher & $395$\,M & $397$\,MB & $19.9$ & $97.6$ & $96.2$ & $97.6$ & $100$ \\
Qwen3Guard-Gen 0.6\,B & $750$\,M & $1.5$\,GB & $33.2$ & $96.2$ & $83.4$ & $44.2$ & $93$ \\
Qwen3Guard-Gen 4\,B   & $4.4$\,B & $8.8$\,GB & $53.7$ & $96.6$ & $86.2$ & $49.1$ & $100$ \\
Qwen3Guard-Gen 8\,B   & $8.2$\,B & $16.4$\,GB & $78.6$ & $96.7$ & $86.8$ & $48.5$ & $100$ \\
\bottomrule
\end{tabular}
\end{table}

\paragraph{Policy-B jury.} Four instruct models (two Gemma-family, two Qwen-family), each sampled
three times at $T{=}0.7$, $\text{top-}p{=}0.95$, so every round carries exactly $12$ votes. The
deployed output holds $144{,}727$ records and $404{,}167$ round instances, with $12$ votes in every
one of them and $1{,}212{,}501$ votes per model; $0.06\%$ of votes failed to parse. The cascade is
three binary rounds: LEGITIMATE/TRUNCATED, then SAFE/UNSAFE, then either
\textsc{boundary\_safe}/\textsc{generic\_safe} or \textsc{compound}/\textsc{harmful}. Cross-model
agreement with the round majority is $94.6\%$ over all valid votes, $95.7\%$ over rounds with an
unambiguous majority, and it splits by round the same way the per-class retention in
\S\ref{sec:setting} does: $98.2\%$ on the UNSAFE branch against $91.6\%$ on the SAFE branch, which is
where \textsc{boundary\_safe} is decided. Rather than force a label, $9{,}431$ tied rounds route
$5{,}632$ records ($3.9\%$) to a separate ambiguous bucket and a further $12{,}191$ are dropped as
truncated, so $12.3\%$ of records never receive a policy-B label. (An earlier draft quoted $85.6\%$
here. That figure is top-1 accuracy for a different, five-model ensemble at the $24$-pillar
labelling stage, measured against a $2{,}400$-sample gold set, and is not jury agreement.)

\paragraph{Arm provenance.} Read from the training configs. All arms use seed $42$ where a seed is
set; \texttt{finetune\_head\_only.py} sets none, never reads the validation split, and selects on
training loss. Its checkpoints nonetheless record \texttt{seed: 42} in their saved config, because
the script writes the base checkpoint's config dictionary through unchanged; the calibration pass
itself is unseeded. E1 and E10 train on the same file (\texttt{calib\_E1\_balanced.jsonl},
$8{,}489$ rows: $5{,}489$ label-disagreement cases plus $3{,}000$ \textsc{generic\_safe} anchors)
and differ only in which checkpoint they initialise from. The explicit policy-refusal anchors are
$1{,}450$ rows of E3's $144{,}744$ ($1.0\%$) and $454$ of E5's $47{,}287$ ($1.0\%$); E2 and E4 carry
none. All $1{,}904$ are drawn from that same $5{,}489$-row disagreement set and all carry policy A's
unsafe label. The $3{,}000$ anchors were added after an earlier calibration set of disagreement cases
only, containing zero \textsc{generic\_safe} examples, erased that class entirely ($526$ of $526$
re-routed to \textsc{boundary\_safe}) and cut $4$-class validation accuracy from $87\%$ to $57.9\%$;
the anchors restored it to $87.9\%$.

\begin{table}[h]
\caption{Arm provenance, read from the training configs.}
\label{tab:prov}
\centering
\small
\begin{tabular}{llrll}
\toprule
Arm & Policy-A route & Pool rows & Initialisation & KD teacher ($149$M only) \\
\midrule
E1  & head-only calibration & $8{,}489$   & policy-B ckpt & n/a (backbone frozen) \\
E10 & head-only calibration & $8{,}489$   & policy-A ckpt & n/a (backbone frozen) \\
E7  & head-only calibration & $53{,}395$  & policy-B ckpt & n/a (backbone frozen) \\
E2  & blend                 & $47{,}287$  & v2 warm-start & policy-B \\
E5  & blend $+$ anchors     & $47{,}287$  & v2 warm-start & policy-B \\
E3  & blend $+$ anchors     & $144{,}744$ & v2 warm-start & policy-B \\
E4  & filter only           & $138{,}638$ & v2 warm-start & policy-B \\
E6  & all hard labels       & $40{,}327$  & v2 warm-start & policy-B \\
E9  & inverse blend         & $40{,}327$  & v2 warm-start & policy-B \\
E8  & teacher only          & $138{,}823$ & v2 warm-start & \textbf{policy-A} \\
\bottomrule
\end{tabular}
\end{table}

At $395$M no arm has a KD teacher, which is why E6 and E9 join the policy-A-last set there and E8
leaves it (\S\ref{sec:install}). The seven full fine-tune arms train for $5$ epochs at $149$M and
$2$ at $395$M; the three head-only arms train for $5$ epochs at both scales, because the orchestrator
hardcodes \texttt{-{}-epochs 5 -{}-lr 5e-5} for every head-only entry. Learning rate is $5$e-$5$ at
$149$M and for every head-only arm, and $3$e-$6$ for the $395$M full arms; the full arms use warmup
fraction $0.06$ and the head-only arms no warmup. \texttt{finetune\_head\_only.py} takes no
validation file and saves on the epoch-average training loss, which in five of the six head-only
runs selected the final epoch, that is, selected nothing. The sixth, E1 at $149$M, stopped after $2$
of its $5$ epochs and was never re-run (its \texttt{best.pt} records
\texttt{epoch\_at\_best}${=}2$); it is therefore the least-trained head-only arm, and it still sits
at the policy-A extreme of both HarmBench-copyright and OR-Bench-hard ($100.0$ and $4.7\%$), so the shortfall
runs against the claim rather than for it.

\paragraph{The rank test, in full.} Equation~\eqref{eq:rank} treats the ten arms as exchangeable
and fixes the extreme end in advance. It does neither well, so we give the ladder.
(i) As printed, one-sided, arms exchangeable: $1/\binom{10}{4}=0.0048$ ($149$M) and
$1/\binom{10}{5}=0.0040$ ($395$M).
(ii) Two-sided, since $\mathcal{A}$ takes the top of HarmBench-copyright and the bottom of
OR-Bench-hard and which end was never fixed: $0.0095$ and $0.0079$.
(iii) Clustering E1, E7 and E10 into one unit and permuting the eight resulting units:
$1/\binom{8}{2}=0.036$ and $1/\binom{8}{3}=0.018$ one-sided, $0.071$ and $0.036$ two-sided. The
clustering is not arbitrary: those three arms are three cells of one $2\times2$,
E1 and E7 sharing a base checkpoint and differing only in the calibration file, E1 and E10 sharing
the calibration file and differing only in the base checkpoint, all three run by one script with no
seed and no validation.
(iv) Dropping the three head-calibration arms altogether, which removes the frozen-backbone and
checkpoint-selection differences with them: $\mathcal{A}=\{$E8$\}$ at $149$M (gaps $+69.3$ and
$+25.0$ points, $p=1/7=0.14$) and $\mathcal{A}=\{$E6, E9$\}$ at $395$M (gaps $+52.0$ and $+12.7$,
$p=1/21=0.048$). The same values follow from clustering under a size-respecting label permutation,
so two defensible clustered conventions differ by a factor of four and we name the one we quote.
The $395$M column survives (iv) and the $149$M column does not.

The four reported cells are not four tests. Spearman across the ten arms between
HarmBench-copyright and OR-Bench-hard is $-0.868$ ($149$M) and $-0.745$ ($395$M), and between the
two scales $+0.716$ and $+0.758$. More sharply, the exact joint permutation (does $\mathcal{A}$
occupy the extreme ranks on \emph{both} benchmarks?) returns exactly the marginal $p$ at every
setting above, because exactly one subset of the right size qualifies on each benchmark and it is
the same subset: the second benchmark contributes no independent evidence. A Bonferroni correction
over four cells ($0.019$ and $0.016$) would therefore claim more tests than were run.

The $395$M grouping was revised after the $149$M definition was applied there and failed: with
$\mathcal{A}=\{$E1, E7, E8, E10$\}$ the $395$M arms hit $3/4$ on HarmBench-copyright and $2/4$ on
OR-Bench-hard. The revision follows from the configs, since no $395$M training config carries a
\texttt{teacher\_ckpt} key and all seven $149$M ones do, so we regard it as principled; it was not
pre-specified, and we read the $395$M $p$ as confirmatory.

\paragraph{The two distillation teachers (E8).} Both are ModernBERT-large warm-started from the
round-3 checkpoint of the same recipe (\texttt{r3b\_warm\_augmented\_drift\_s42}, each at its own
data version) at seed $42$, $2$ epochs, learning rate $3\times10^{-6}$, so lineage, backbone and
schedule are matched. Data volume is not. The policy-A teacher
(\texttt{r4b\_warm\_r3b\_s42\_v3\_5}, on \texttt{round3\_train\_r4\_augmented\_clean.jsonl}) saw
$67{,}352$ unique prompts after normalisation; the policy-B teacher
(\texttt{r4b\_warm\_r3b\_s42\_v10}, on \texttt{train\_v10\_balanced.jsonl}) saw $138{,}805$, a factor
of $2.06$, and $107{,}044$ of those ($77.1\%$ of that pool, $32{,}715$ of them harmful-class) never
appear in the policy-A teacher's pool. E8's contrast is a teacher swap, not a single-variable one,
though the confound runs against the effect: the teacher that installs policy A is the one trained
on less data.

\paragraph{Pool size and route.} At $395$M the two are collinear: Spearman
$\rho(\text{pool},\text{route})=-0.738$ over the ten arms and $-0.805$ among the seven full
fine-tunes, giving $\rho(\text{pool},\text{OR-Bench-hard})=+0.917$ and
$\rho(\text{pool},\text{HarmBench-copyright})=-0.654$. At $149$M it is absent ($-0.215$; $+0.416$
among the full fine-tunes). Route reproduces the extreme-$k$ grouping exactly at both scales and pool
size does not: the five smallest pools are $\{$E1, E10, E2, E6, E9$\}$ against an observed extreme
five of $\{$E1, E10, E6, E7, E9$\}$, E7 ($53{,}395$ rows) sits on the policy-A side, and at $149$M
the four policy-A-last arms span $8{,}489$--$138{,}823$ rows inside a $3.4$-point over-refusal band.
The honest caveat: \emph{within} the mixing arms pool size does track over-refusal ($\rho=+0.971$ at
$149$M, $n{=}6$; $+0.912$ at $395$M), so the OR-Bench-hard magnitudes, unlike the grouping, are
partly a pool effect.

\paragraph{Jury agreement.} The relabelling jury's per-model votes were retained, so a
chance-corrected statistic is computable and we compute one. Over each jury model's modal vote,
Fleiss $\kappa=0.78$ on the binary safe/unsafe round ($N{=}132{,}470$ items; $0.82$ treating every
vote as a separate rater), $0.70$ on the \textsc{boundary\_safe} versus \textsc{generic\_safe}
refinement ($N{=}55{,}381$) and $0.87$ on \textsc{compound} versus \textsc{harmful}
($N{=}71{,}507$). Within-model self-consistency across three samples is $93.6$--$99.7\%$ and pairwise
raw agreement between models $84.1$--$94.4\%$, so the residual disagreement is between models rather
than sampling noise. Agreement is lowest on exactly the contested class this paper is about, and
none of it speaks to correctness.

\paragraph{Prompt-sampling dispersion in $\Delta_b$.} Resampling prompts with replacement inside each
benchmark and rescoring all ten arms at both scales on the same resample ($B{=}2{,}000$) gives
$\mathrm{SD}(\Delta_b)$ between $0.26$ (SaladBench-base) and $3.39$ points (HarmBench-contextual),
with $95\%$ intervals of $99.3\,[98.0,100.0]$ on HarmBench-copyright, $57.4\,[52.8,62.9]$ on
OR-Bench-hard, $33.4\,[30.6,36.4]$ on DoNotAnswer and $4.3\,[3.5,5.4]$ on WildGuard-benign. What
makes the middle of Figure~\ref{fig:audits}(b) unreadable is therefore seed variance, not prompt
sampling. Two cautions: $\max-\min$ is biased upward under resampling, by $0.0$--$2.0$ points here
and worst on the $100$-prompt subsets; and the arms are scored on byte-identical prompt lists, so an
independent-binomial floor would be the wrong null. Under the matched test every one of the $28$
benchmark $\times$ scale extreme-arm gaps is significant, worst case exact McNemar $p=0.008$.

\paragraph{Per-arm benchmark exposure.} Share of OR-Bench-hard's $1{,}319$ prompts present in each
arm's own training or calibration pool: E9 $1.4\%$, E8 $37.0\%$, E4 $38.7\%$, E3 $39.3\%$,
E1/E10 $40.3\%$, E6 $63.2\%$, E2/E5/E7 $73.8\%$. Rank correlation with the \textsc{OR-h} column of
Table~\ref{tab:main} is $-0.01$ ($149$M) and $-0.21$ ($395$M), computed on the percentages as printed
here; the E2/E5/E7 tie is an artefact of that rounding, and with the unrounded overlaps ($974$, $973$,
$973$ prompts) the $149$M figure is $0.00$. Effective exposure is far flatter than these own-pool
numbers. Following \texttt{init\_from} and \texttt{teacher\_ckpt} through every training config, each
arm also inherits the pools of the runs it was warm-started or distilled from, and all twenty lineages
reach \texttt{r2a\_t6\_everything\_s42} (\texttt{train\_p3c\_balanced\_filt500.jsonl}) and then
\texttt{p2\_c\_pillar\_only} (the original \texttt{train.jsonl}/\texttt{val.jsonl} split); those three
files alone cover $81.0\%$ of OR-Bench-hard. Over the whole traced lineage, effective exposure is
$82.3\%$ for seventeen of the twenty models, $81.7\%$ for E8 at $149$M (no warm start, only a
\texttt{v3\_5} KD teacher) and $81.0\%$ for the two E10 arms (calibrated on \texttt{v3\_5} bases),
against the $82.1\%$ of the benchmark that the union of all pools contaminates. Every other arm
inherits through the \texttt{v2} relabelling runs, so it is the shared ancestry, not any single
intermediate pool, that flattens exposure. Union decontamination also costs nothing on this
benchmark: no OR-Bench-hard prompt is removed only because of the policy-A head-calibration pools,
and total contamination is $1{,}083/1{,}319$ either way.

\paragraph{Validation-split exposure.} Share of the $7{,}658$-prompt validation split present in each
arm's own training pool: E1/E10 $10.9\%$, E2 $23.2\%$, E5 $24.2\%$, E8 $35.9\%$, E3 $36.8\%$,
E4 $37.0\%$, E9 $40.0\%$, E6 $52.8\%$, E7 $69.3\%$. Rank correlation with the \textsc{Val} column is
$-0.15$ ($149$M) and $+0.16$ ($395$M).

\section{Prior art on the contested middle class, and on contamination}
\label{app:priorart}

Qwen3Guard~\citep{qwen2025qwen3guard} is the closest prior work to \S\ref{sec:sota}. Its \S3.4.1
reports that labelling \emph{Controversial} samples unsafe better matches Aegis while labelling them
safe better matches OpenAIModeration; its \S3.4.2 (Table~8, Qwen3Guard-Gen-$4$B only) ablates the
class; and the captions of its results tables declare a per-benchmark optimal mode. Its \S3.3 also
reports a training-time effect: relative to a $1{:}1$ Safe-to-Unsafe ratio, doubling the proportion
of Safe examples makes the model more permissive, and it uses two such re-weighted models'
disagreement to construct the \emph{Controversial} label. Two things separate that from our claim.
First, there the lever is the class \emph{ratio} inside one fixed taxonomy, used as
label-construction machinery, and its effect is never measured on public safety benchmarks; here the
lever is the \emph{definition}, and we measure it across fifteen decontaminated benchmarks and twenty
models. Second, its distillation stage is motivated as annotation-noise reduction and makes no
policy-transfer claim, whereas we find distillation carries the teacher's policy over contrary hard
labels. On contamination, the closest guardrail-specific work is \citet{young2025guardrails}, which
evaluates ten released guards on $1{,}445$ prompts across $21$ attack categories and finds every one
degrades on novel attacks, the largest gap being Qwen3Guard-$8$B's fall from $91.0\%$ to $33.8\%$ and
the smallest Granite-Guardian-3.2-$5$B's $6.5$ points; that inference is behavioural, where
\S\ref{sec:contam} measures corpus overlap directly. On taxonomy adaptation, given a $25$-category
in-house taxonomy at inference LlamaGuard3-$8$B scores $3.5\%$ F1 against $84.9\%$ on
HarmBench~\citep{roblox2024guard}.

\section{Seed floors: all $37$ triplets}
\label{app:floor}

A \emph{triplet} is one recipe trained at seeds $42/43/44$ with configs identical apart from
\texttt{run\_id} and \texttt{seed} (with ten exceptions noted below), scored on the fifteen-benchmark
suite of Appendix~\ref{app:suite}. The training tree holds $37$ complete triplets, $111$ runs;
they are nine lineages at up to six data versions each, not $37$ independent recipes. The
four used in \S\ref{sec:sens} are the \texttt{v10x} family, whose configs differ from arm E3's in
exactly one key, \texttt{binary\_mode} (a $2$-class rather than $4$-class head), plus the training
pool: initialisation, learning rate, epochs, backbone, class and pillar loss weights ($0.01$ and
$1.0$) and checkpoint-selection metric (pillar macro-F1, the default, which no config in the grid or
the triplets overrides) are those of the $395$M arms. In both families the benchmark verdict is read
off the class head (\texttt{eval\_external\_benchmarks.py}), so at $395$M the floor recipes' head is
supervised and selected exactly as the arms' is. At $149$M the arms use a class-loss weight of $1.0$
and a smaller backbone, which is why we read the floor as off-scale there.

Table~\ref{tab:floor} gives $\Phi_b$ over all $37$ triplets against the value used in
\S\ref{sec:sens}, with $\Delta_b/\Phi_b$ under each. Three things follow.

\begin{table}[!ht]
\caption{Seed floors on the suite of Appendix~\ref{app:suite}, in points. $\Delta_b$ is the
label-supervision range of \eqref{eq:delta} on decontaminated subsets. The three $\Phi_b$ columns are
the minimum, median and maximum seed spread over all $37$ triplets; \textsc{used} is the median over
the four \texttt{v10x} recipes quoted in \S\ref{sec:sens}, the family that supplies the maximum of the
$37$ on the first two rows and on SaladBench-attack and WildGuard-benign. The last two columns give
$\Delta_b/\Phi_b$ under that floor and under the median of all $37$, both computed before rounding.}
\label{tab:floor}
\centering
\small
\begin{tabular}{lr rrr r rr}
\toprule
& & \multicolumn{3}{c}{$\Phi_b$ over all $37$ triplets} & & \multicolumn{2}{c}{$\Delta_b/\Phi_b$} \\
\cmidrule(lr){3-5}\cmidrule(lr){7-8}
Benchmark & $\Delta_b$ & min & median & max & \textsc{used} & used & all-$37$ \\
\midrule
HarmBench-copyright & $99.3$ & $0.0$ & $4.0$ & $71.0$ & $58.5$ & $1.70$ & $24.83$ \\
OR-Bench-hard & $57.4$ & $0.2$ & $2.8$ & $58.1$ & $38.6$ & $1.49$ & $20.47$ \\
OR-Bench-toxic & $38.5$ & $0.6$ & $3.1$ & $28.9$ & $18.2$ & $2.12$ & $12.60$ \\
DoNotAnswer & $33.4$ & $0.1$ & $2.0$ & $18.8$ & $3.9$ & $8.46$ & $16.48$ \\
SaladBench-defense & $31.0$ & $0.5$ & $4.5$ & $39.5$ & $7.0$ & $4.43$ & $6.89$ \\
HarmBench-contextual & $29.0$ & $1.0$ & $5.0$ & $14.0$ & $10.0$ & $2.90$ & $5.80$ \\
SaladBench-attack & $25.2$ & $0.3$ & $3.4$ & $42.2$ & $37.0$ & $0.68$ & $7.51$ \\
SaladBench-base & $21.7$ & $0.2$ & $1.6$ & $32.4$ & $11.1$ & $1.96$ & $13.58$ \\
AdvBench & $19.8$ & $0.2$ & $1.2$ & $50.2$ & $12.6$ & $1.57$ & $16.63$ \\
StrongREJECT & $15.5$ & $0.3$ & $1.0$ & $44.7$ & $6.1$ & $2.55$ & $16.12$ \\
HarmBench-standard & $14.1$ & $0.0$ & $1.5$ & $53.0$ & $5.8$ & $2.45$ & $9.39$ \\
JailbreakBench-harmful & $13.1$ & $0.0$ & $2.0$ & $49.0$ & $7.0$ & $1.87$ & $6.53$ \\
WildGuard-harmful & $8.9$ & $0.4$ & $1.5$ & $21.1$ & $5.9$ & $1.51$ & $6.10$ \\
WildGuard-benign & $4.3$ & $0.1$ & $0.6$ & $10.7$ & $5.0$ & $0.85$ & $6.73$ \\
\bottomrule
\end{tabular}
\end{table}

\textbf{The floor is a recipe-level choice, not a property of the benchmark.} On HarmBench-copyright
the median spread is $4.0$ points over all $37$ triplets and $3.0$ over the $33$ that are not
\texttt{v10x}, against $58.5$ over the four we use: a factor of $19.5$ between the family we chose and
the rest, and $24.2$ on OR-Bench-hard ($38.6$ against $1.6$).

\textbf{The choice runs against our own claims.} The widest of the $37$ triplets on
HarmBench-copyright, OR-Bench-hard, SaladBench-attack and WildGuard-benign is in every case a
\texttt{v10x} recipe, and those are exactly the four benchmarks \S\ref{sec:sens} and
\S\ref{sec:install} turn on.

\textbf{The ordering by $\Delta_b/\Phi_b$ is not identified by the triplets available.} Under the
floor we use, SaladBench-attack ($0.68\times$) and WildGuard-benign ($0.85\times$) fall below $1$.
Under the median of all $37$, of the other $33$, or of the two on-path triplets named at the end of
this appendix, nothing falls below $1$, and those three orderings correlate with the printed one at
Spearman $-0.03$, $-0.08$ and $+0.26$. The ratio also stops re-ranking: its correlation with raw
$\Delta_b$ rises from $+0.24$ under our floor to $+0.58$ and $+0.59$ under the first two. We claim the
top/bottom split, not the order inside it.

\paragraph{Levels, and what the level claim is scoped to.} \S\ref{sec:sens} compares arm levels to
what the twelve \texttt{v10x} runs ever reached, on the full benchmarks, since that is where $\Phi_b$
is measured. On HarmBench-copyright those runs span $2$--$73$ and the policy-A-last arms sit at $100$
($149$M, all four) and $74$--$100$ ($395$M); on OR-Bench-hard the runs span $11.2$--$72.3\%$ and the
arms $1.6$--$3.3\%$ and $2.5$--$3.8\%$. At $395$M the copyright margin is one point ($74$ against
$73$), and replacing the arm levels by their decontaminated values removes it ($66.7$ for E9); the
same substitution puts two $395$M arms above the OR-Bench-hard minimum ($12.7\%$ for E7, $11.4\%$ for
E9). The $149$M comparison holds either way ($100$ decontaminated as well, and $3.4$--$6.8\%$).
The claim is also scoped to those four recipes: across all $111$ runs of the $37$ triplets
HarmBench-copyright reaches $100$ and OR-Bench-hard falls to $0.9\%$, but those runs differ in their
label supervision, which is the effect $\Delta_b$ measures, not in their seed.

\paragraph{Caveats on the $37$.} Ten are not strictly seed-only: five have an \texttt{init\_from} that
differs between seeds, and five are head-only finetunes stacked on those five. Restricting to the $27$
strict triplets changes little: HarmBench-copyright min/median/max $0.0/6.0/71.0$, OR-Bench-hard
$0.2/3.0/58.1$, and the non-\texttt{v10x} strict medians are $4.0$ and $1.5$. Separately, the four
recipes we use are not four independent draws on HarmBench-copyright. All four are monotone increasing
in the seed index ($2/13/57$, $2/5/73$, $3/23/57$, $3/26/65$), which has probability
$(1/6)^4 = 7.7\times 10^{-4}$ if seeds are exchangeable, so on that benchmark the seed index behaves
like a systematic factor rather than like noise, and the median of the four spreads is closer to one
draw than to twelve. The pattern is specific to it: OR-Bench-hard, OR-Bench-toxic and
SaladBench-attack are $0/4$ monotone, and only WildGuard-benign ($3/4$, all decreasing) comes close.

\paragraph{Why the floor is off-path.} None of the twenty arms has a seed sibling: each was trained
once at seed $42$, so $\Phi_b$ for the arms' own recipes is not measurable from these artefacts at
all, which is why a proxy is used. The nearest on-path triplets are the \texttt{v2} warm-start
checkpoint the seven fully retrained arms initialise from (HarmBench-copyright spread $3.0$ points,
OR-Bench-hard $1.5$) and the $149$M distillation teacher of E2--E6 and E9 ($2.0$ and $6.9$), both far
below the floor we quote; E1, E7 and E10 have frozen backbones and start from later checkpoints, of
which only the $395$M E1/E7 base has a triplet at all, and it is one of the ten flagged above ($7.0$
and $3.8$). A lineage name alone does not identify a floor: the variant of the warm-start recipe the
arms actually initialise from has a HarmBench-copyright spread of $3.0$ points, while the same recipe
on the earlier data version has $12.0$. Closing this properly needs $40$ further training runs, two
extra seeds per arm.

\paragraph{Prompt sets.} $\Phi_b$ is measured on the full benchmarks and $\Delta_b$ on the
decontaminated subsets, so the ratio is indicative rather than exact. Recomputing $\Delta_b$ on the
full benchmarks moves it by under $5\%$ on twelve of the fourteen (median $1.5\%$), by $11.3\%$ on
WildGuard-benign ($4.3 \to 4.8$) and by $36.6\%$ on OR-Bench-hard ($57.4 \to 42.0$), which takes that
benchmark's ratio from $1.49$ to $1.09$. Both entries below $1$ survive the swap: SaladBench-attack
$0.68$ either way, WildGuard-benign $0.85 \to 0.96$.

\section{The over-refusal label audit}
\label{app:jury}

\paragraph{Instrument.} Every one of the $20{,}456$ prompts harvested for the \textsc{boundary\_safe}
class was labelled independently by four instruction-tuned models of $26$--$35$B parameters
(two Gemma-family, two Qwen-family) at temperature $0$, into \{\textsc{boundary\_safe},
\textsc{generic\_safe}, \textsc{harmful}, \textsc{ambiguous}\}, from a two-factor prompt (does it
\emph{look} harmful, and would a complete answer \emph{do} harm). The \textsc{harmful} bar is written
strictly, so the counts below are conservative. The jury ran once, with no human adjudication and no
inter-rater statistic beyond the vote counts, so it is our instrument and not ground truth.

\paragraph{Pool-level split.} At a uniform $3$-of-$4$ bar: $50.9\%$ \textsc{boundary\_safe}, $32.9\%$
\textsc{generic\_safe}, $4.5\%$ \textsc{harmful}, $11.6\%$ with no three-vote majority. By plurality,
which is a partition: $60.3$ / $33.5$ / $6.3\%$, with \textsc{ambiguous} never winning a single row of
the $20{,}456$. At the looser two-of-four bar, $6.9\%$ of the pool is \textsc{harmful}; by source that
is $17.9\%$ of the OR-Bench-hard prompts and $26.4\%$ of the FalseReject prompts.

\paragraph{The clean $236$.} Vote patterns, covering all $236$: $92\times$\{bs $2$, harmful $2$\},
$72\times$\{bs $1$, harmful $3$\}, $39\times$\{harmful $4$\}, $25\times$\{bs $2$, gs $1$, harmful
$1$\}, $7\times$\{bs $1$, gs $1$, harmful $2$\}, $1\times$\{gs $1$, harmful $3$\}. So $211$ ($89.4\%$)
are flagged harmful by two or more jurors and $112$ ($47.5\%$) by three or more ($39$ unanimously),
none reaches three \textsc{boundary\_safe} votes, and $92$ are exact $2$--$2$ ties, better read as
contested than as resolved. Against the full benchmark, $966$ of the $1{,}319$ are jury-confirmed and
all $966$ are contaminated; the $353$ the jury rejected split $236$ clean and $117$ contaminated (the
repository also contains passes that promote flagged and unclear prompts back into the class). Mean
flag rate over the twenty arms is $0.160$ on the $1{,}083$ contaminated prompts, $0.302$ on the clean
$236$, $0.344$ on the $112$ and $0.110$ on the $25$; the label-supervision range is $0.392$, $0.574$,
$0.643$ and $0.260$ on the same four sets. Composition explains the first series but not the second.

\paragraph{Joining the audit.} Only $1{,}221$ audit rows carry \texttt{source=orbench\_hard}: the pool
builder adds OR-Bench-$80$k before OR-Bench-hard and deduplicates on a global hash of the text, so $98$
OR-Bench-hard prompts are stored under the other tag. Joining on normalised text instead recovers all
$1{,}319$, so the audit covers the benchmark completely and none of the figures above is a coverage
artefact.

\paragraph{Stratified over-refusal (\S\ref{sec:sota}).} OR-Bench-hard over-refusal on the clean $236$,
then on the $211$ the jury flags harmful at two or more votes, the $112$ at three or more, and the $25$
it leaves unclear: ours $149$M (E8) $6.4$ / $6.6$ / $9.8$ / $4.0$; ours $395$M (E10) $10.2$ / $10.9$ /
$11.6$ / $4.0$; Qwen3Guard-$0.6$B $56.4$ / $61.1$ / $63.4$ / $16.0$; Qwen3Guard-$4$B $57.6$ / $64.0$ /
$63.4$ / $4.0$; Qwen3Guard-$8$B $55.9$ / $61.1$ / $62.5$ / $12.0$; Granite Guardian $97.0$ / $98.6$ /
$98.2$ / $84.0$. On the last stratum, the only one that resembles a real over-refusal test, the
Qwen3Guard gap does not survive the label audit; the Granite gap does. $n{=}25$, so read it as a
direction and not an estimate.

\end{document}